# HistoPerm: A Permutation-Based View Generation Approach for Improving Histopathologic Feature Representation Learning

**Short running title: HistoPerm**

Joseph DiPalma, BS[1], Lorenzo Torresani, PhD[1], Saeed Hassanpour, PhD[1,2,3*]

[1]Department of Computer Science, Dartmouth College, Hanover, NH 03755, USA
[2]Department of Biomedical Data Science, Geisel School of Medicine at Dartmouth, Hanover, NH 03755, USA
[3]Department of Epidemiology, Geisel School of Medicine at Dartmouth, Hanover, NH 03755, USA

[*]Corresponding Author: Saeed Hassanpour, PhD
Postal address: One Medical Center Drive, HB 7261, Lebanon, NH 03756, USA
Email: Saeed.Hassanpour@dartmouth.edu

# Abstract

Deep learning has been effective for histology image analysis in digital pathology. However, many current deep learning approaches require large, strongly- or weakly-labeled images and regions of interest, which can be time-consuming and resource-intensive to obtain. To address this challenge, we present HistoPerm, a view generation method for representation learning using joint embedding architectures that enhances representation learning for histology images. HistoPerm permutes augmented views of patches extracted from whole-slide histology images to improve classification performance. We evaluated the effectiveness of HistoPerm on two histology image datasets for Celiac disease and Renal Cell Carcinoma, using three widely used joint embedding architecture-based representation learning methods: BYOL, SimCLR, and VICReg. Our results show that HistoPerm consistently improves patch- and slide-level classification performance in terms of accuracy, F1-score, and AUC. Specifically, for patch-level classification accuracy on the Celiac disease dataset, HistoPerm boosts BYOL and VICReg by 8% and SimCLR by 3%. On the Renal Cell Carcinoma dataset, patch-level classification accuracy is increased by 2% for BYOL and VICReg, and by 1% for SimCLR. In addition, on the Celiac disease dataset, models with HistoPerm outperform the fully-supervised baseline model by 6%, 5%, and 2% for BYOL, SimCLR, and VICReg, respectively. For the Renal Cell Carcinoma dataset, HistoPerm lowers the classification accuracy gap for the models up to 10% relative to the fully-supervised baseline. These findings suggest that HistoPerm can be a valuable tool for improving representation learning of histopathology features when access to labeled data is limited and can lead to whole-slide classification results that are comparable to or superior to fully-supervised methods.

# 1. INTRODUCTION

Digital pathology involves the visualization and analysis of whole-slide images (WSIs) to assist pathologists in the diagnosis and prognosis of various diseases. These WSIs are digitized at high resolutions and can be analyzed manually, using computer vision models, or a combination. However, the large size of these images, up to 150,000×150,000 pixels, can present challenges for typical computer vision-based image analysis tools.

In recent years, various computer vision-based methods and solutions have been proposed and developed to handle the gigapixel size of WSIs and address other unique challenges of digital pathology.[1–8] In terms of label and annotation requirements, these methods differ from those used on natural images in several ways. Firstly, the labeling process for WSIs requires highly trained experts, while natural images often require minimal or no prerequisites for labeling. Secondly, labels are typically provided at the slide level rather than at the patch level. Finally, the class label may only be determined by a small portion of the WSI. These characteristics present major challenges for the application of standard computer vision methods in digital pathology.

Among these three annotation bottlenecks, the last two are most unique to digital pathology. Due to the large size of the WSIs, it is infeasible for pathologists to label all regions of interest on a slide. Instead, the label is usually provided at the slide level, which also applies to class-negative regions of a slide. Moreover, an object in the average image from the ImageNet natural image dataset occupies 25% of the area,[9] while a typical region-of-interest annotation in a WSI can occupy as little as 5% of the image.[4,10] The combination of weak labeling and low object scale poses a unique challenge and makes applying standard computer vision methods a suboptimal solution in digital pathology.

In the last decade, deep learning models have been highly successful in numerous classic computer vision tasks.[11–13] To make these standard deep learning models more feasible and effective for WSIs, it is common to preprocess the images into smaller patches, typically 224×224 pixels. However, this can lead to further issues if the weakly labeled nature of the slides is not considered. A common approach involves the use of a convolutional neural network (CNN) on smaller patches extracted from large whole-slide images, with the patch classification results being aggregated for whole-slide inferencing.[14–21] However, this approach can have suboptimal performance if the signal-to-noise ratio is low among the extracted patches. More advanced methods involving attention[22,23] or multiple-instance learning[24–32] have been developed to use the weak-labels, but these still require large, labeled datasets.

In recent years, self-supervised representation learning techniques have gained significant traction for their ability to solve difficult problems in computer vision without relying on labor-intensive, manually-labeled datasets. These methods utilize a pretext task to learn a latent representation of an unlabeled dataset, which is often readily available in the medical domain. To address the labeling challenge, self-supervised approaches have been successfully applied to histology images using existing computer vision techniques.[33–37] These approaches aim to exploit the unique characteristics of these images, such as rotation invariance or local-to-global consistency. In addition, contrastive learning-based methods have gained popularity in histology feature representation,[38–45] with techniques such as Contrastive Predictive Coding[38,42] and DSMIL[46] proving effective for incorporating multiscale information into contrastive models. However, all these approaches still require all input data to be labeled, whether weakly or strongly.

To address this shortcoming, we propose a model-agnostic view generation method called HistoPerm for representation learning in histology image classification. Unlike prior methods, HistoPerm is flexible and incorporates both labeled and unlabeled data into the learning process. In contrast to prior view generation approaches for histology images, which produce views at random from the same instance, we perform a permutation on a portion of the mini-batch such that the view comes from the same class but a different instance of the class. By taking advantage of the large pool of both class-positive and class-negative patches, our approach can derive stronger representations for histological features. Our experiments show that adding HistoPerm to an existing state-of-the-art representation learning image analysis pipeline improves the histology image classification performance.

## 1.1 Representation Learning with Joint Embedding Architectures

Several paradigms for representation learning have been proposed, including contrastive learning,[47–53] noncontrastive learning,[54–56] and information preservation.[57–60] These paradigms all employ joint embedding architectures, where two models are trained to produce similar outputs when given augmented views of the same source image.

Contrastive learning relies on positive and negative samples to guide the network through learning unique identifiers for each class in the downstream task. However, these approaches often require large mini-batch sizes and significant computational resources, making them impractical for many studies and applications. Smaller mini-batch sizes can be used by implementing "tricks" such as momentum encoders,[49,50] but in general, these approaches are still resource intensive and require massive computational power that is inaccessible to most researchers.

Noncontrastive approaches, which utilize only positive instances, also require fewer resources but may result in a slight decrease in the downstream classification performance.[54–56] The underlying principle that prevents convergence to trivial, constant (i.e., collapsed) embeddings in these methods is unknown, but prior works have shown that implementation details do play some part in their success.[61–63]

Information preservation methods, such as Barlow Twins,[64,65] Whitening-MSE,[65] and VICReg,[66,67] aim to decorrelate variables in the learned representations and explicitly prevent collapse. These methods are effective at avoiding trivial embeddings and have shown promise in natural image tasks.

**1.2 Rationale for Our Work**

Prior work has shown that representation learning methods rely on building representations that are invariant to irrelevant variations in the input.[68] For histopathology, many patches share similar histologic features and visual attributes, independent of the class. Given this, many of the patches sampled from WSIs are unsuitable as negative samples for learning unlike natural images. Hence, we utilize the large pool of *both* class-positive and class-negative patches to build stronger representations for histologic features by allowing permutation at the mini-batch scale. While we may encounter instances where class-positive and class-negative instances are paired and these instances are *not* morphologically similar, these hard cases should not be common enough in a typical WSI classification task to impact feature learning adversely, and such hard cases may even be beneficial to learning according to previous research.[69] Notably, HistoPerm is model-agnostic and can be integrated into any joint-embedding architecture-based representation learning framework operating on two input views to improve histologic feature representation learning.

## 2. METHOD

In this section, we introduce our proposed method, HistoPerm, a permutation-based view generation technique to improve capturing histologic features in representation learning frameworks. A high-level overview of our approach is shown in Figure 1.

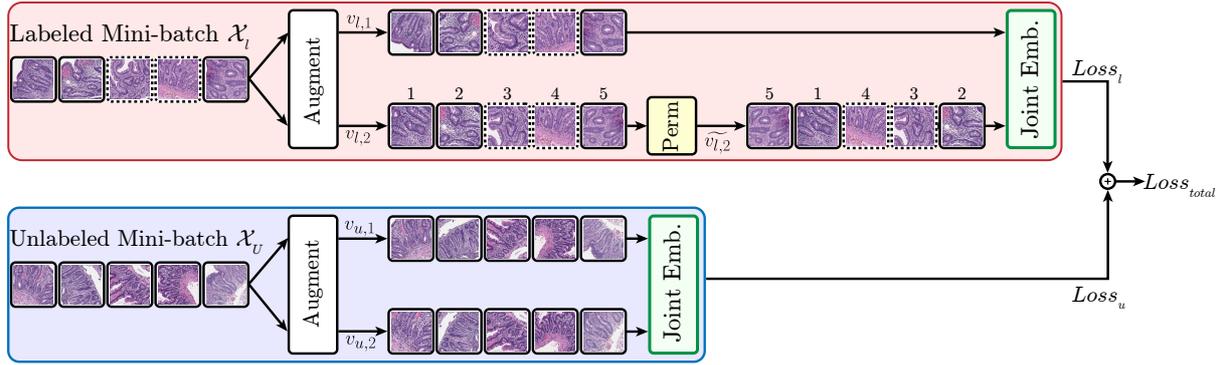

**Figure 1.** Overview of our HistoPerm method. The joint embedding networks are fed randomly augmented views $v_{*,1}$ and $v_{*,2}$. For the labeled mini-batch of patches, $\mathcal{X}_l$, the solid or dashed patch outlines represent the labels. The numbers for labeled views $v_{l,2}$ and $\widetilde{v_{l,2}}$ show the change in patch order before and after the permutation operation, respectively. The unlabeled mini-batch of views, $\mathcal{X}_u$, is fed to joint embedding networks without permutation as in the standard architectures.

## 2.1 WSI to Patch Conversion

Let $\mathcal{D}$ be a dataset comprised of WSIs. Disjoint labeled and unlabeled subsets $\mathcal{D}_l$ and $\mathcal{D}_u$ partition $\mathcal{D}$. For each WSI $\mathcal{S}_i \in \mathcal{D}_l$, we have an associated ground-truth label $y_i$ corresponding to the pathologist-provided slide-level classification label. Given a slide $\mathcal{S}_i$, we produce a set of patches $p_i$ and assign them the slide-level label $y_i$. In this weakly labeled setting, we can have anywhere between 1 and $|p_i|$ class-positive patches per set $p_i$. As discussed earlier, in histopathological classification, we can assume that the majority of patches in $p_i$ will be negative relative to slide-level class $y_i$. After this step, we have labeled and unlabeled sets of patches $\mathcal{P}_l$ and $\mathcal{P}_u$ produced from $\mathcal{D}_l$ and $\mathcal{D}_u$, respectively. In the next section, we explain how we generate the input views for our model, given $\mathcal{P}_l$ and $\mathcal{P}_u$.

## 2.2 View Generation

**View Augmentation.** Given a weakly-labeled patch dataset $\mathcal{P}_l$ and unlabeled patch dataset $\mathcal{P}_u$, we sample mini-batches $\mathcal{X}_l$ and $\mathcal{X}_u$ such that $\mathcal{X}_l \sim \mathcal{P}_l$ and $\mathcal{X}_u \sim \mathcal{P}_u$. Furthermore, let $\alpha \in [0, 1]$ be a hyperparameter representing the fraction of the mini-batch sampled from $\mathcal{P}_l$. Given a mini-batch size of $N$, we have $|\mathcal{X}_l| = \lfloor \alpha \cdot N \rfloor$ and $|\mathcal{X}_u| = N - |\mathcal{X}_l|$. When $\alpha = 0$, this reduces to the default view generation scheme. Starting with data transformation sets $\mathcal{T}_1$

and $\mathcal{T}_2$, we compute $x_{u,i}^{(1)} = t_1(x_{u,i})$ and $x_{u,i}^{(2)} = t_2(x_{u,i})$ with $t_1 \sim \mathcal{T}_1$ and $t_2 \sim \mathcal{T}_2$ for all $x_{u,i} \in \mathcal{X}_u$. These augmented instances are combined to form views $v_{u,1} = \{x_{u,1}^{(1)}, x_{u,2}^{(1)}, \ldots, x_{u,|\mathcal{X}_u|}^{(1)}\}$ and $v_{u,2} = \{x_{u,1}^{(2)}, x_{u,2}^{(2)}, \ldots, x_{u,|\mathcal{X}_u|}^{(2)}\}$. Analogously, we produce augmented views of the labeled mini-batch $\mathcal{X}_l$, denoted as $v_{l,1} = \{x_{l,1}^{(1)}, x_{l,2}^{(1)}, \ldots, x_{l,|\mathcal{X}_l|}^{(1)}\}$ and $v_{l,2} = \{x_{l,1}^{(2)}, x_{l,2}^{(2)}, \ldots, x_{l,|\mathcal{X}_l|}^{(2)}\}$. Next, we describe the view permutation process on labeled views $v_{l,1}$ and $v_{l,2}$.

**View Permutation.** Given labeled augmented views, $v_{l,1}$ and $v_{l,2}$, we define a bijective permutation function $\pi: \{1, \ldots, |\mathcal{X}_l|\} \to \{1, \ldots, |\mathcal{X}_l|\}$ to generate a random permutation of $v_{l,2}$ denoted as $\widetilde{v_{l,2}}$. Our permuted view, $\widetilde{v_{l,2}}$, is defined as $\widetilde{v_{l,2}} = \{x_{l,\pi(i)}^{(2)}: x_{l,i}^{(2)} \in v_{l,2}, y_i = y_{\pi(i)}\}$. Now, $\widetilde{v_{l,2}}$ is a permutation of $v_{l,2}$ where the original image differs, but the ground-truth class is the same. Through this permutation, we augment the size of possible view pairings, enabling the model to learn richer representations. Note that it was an arbitrary choice to permute $v_{l,2}$, and due to symmetry either view could be permuted without loss of generality.

## 3. EXPERIMENTAL SETUP

### 3.1 Datasets

We applied and evaluated our approach on two datasets from an academic medical center. Our datasets are representative of Celiac Disease (CD) and Renal Cell Carcinoma (RCC). Each dataset consists of hematoxylin-eosin-stained, formalin-fixed, paraffin-embedded slides scanned at either 20× (0.5 μm/pixel) or 40× (0.25 μm/pixel) magnification. For run-time purposes, we downsampled the slides to 5× (2 μm/pixel) magnification using the Lanczos filter.[70] We divided the slides into overlapping 224×224-pixel patches for use with the PyTorch deep learning framework.[71] A different overlapping factor was used across each class in the training set to produce approximately 80,000 patches per class. For the development and testing sets, we used a constant overlap factor of 112 pixels among patches. We provide dataset statistics in the supplementary material. Although these datasets are labeled in their original form, we have ignored the labels for a portion of the dataset used in the unlabeled section of the architecture in each epoch according to the formulation provided in Section 2.1 to simulate the intended use of our approach. This means that $\mathcal{D}_l$ and $\mathcal{D}_u$ are varying and built dynamically in each epoch.

## 3.2 Implementation Details

**Image Augmentation.** We used a typical set of image augmentations in our experiments according to common joint embedding architecture-based representation learning methods. A crop from each image is randomly selected and resized to 224×224 pixels with bilinear interpolation. Next, we randomly flip the patches over both the horizontal and vertical axes, as histology patches are rotation invariant. Finally, we performed random Gaussian blurring on the augmented images. Empirical justification, as well as exact implementation details for these transformations, are provided in the supplementary material.

**Pretraining.** In the pretraining phase, we used the LARS optimizer[74] for 50 epochs of training the networks with a 5-epoch warm-up and cosine learning rate decay[75] thereafter. The initial learning rate was 0.45 with a mini-batch size of 256 and weight decay of $10^{-6}$. We choose $\alpha = 0.75$ (i.e., 64 unlabeled and 192 labeled examples) as the optimal balance between the unlabeled and labeled portions of the mini-batch. We provide details of how we selected $\alpha = 0.75$ in the supplementary material. For experiments without HistoPerm, all 256 examples in the mini-batch are considered unlabeled.

**Linear Evaluation.** Linear training uses the SGD optimizer with Nesterov momentum[76] for 80 epochs of training a linear layer on top of the frozen encoders with cosine learning rate decay.[75] We used an initial learning rate of 0.2 and a mini-batch size of 256. Unlike the pretraining step, we only performed affine transformations to the input data. In this phase, we utilized all data in the respective training set.

## 4. RESULTS

### 4.1 Patch-level Results

First, we investigated the effect of HistoPerm on model patch-level classification performance. Table 1 shows that the use of HistoPerm consistently resulted in improved accuracy compared to baseline approaches across all datasets. Specifically, BYOL with HistoPerm outperformed standard BYOL by 8% and 2% on the CD and RCC datasets, respectively, in terms of classification accuracy. SimCLR with HistoPerm also demonstrated improved accuracy by 3% and 1% on the CD and RCC datasets, respectively. Additionally, the incorporation of HistoPerm into VICReg led to an increase in accuracy by 8% and 2% on the CD and RCC datasets, respectively.

| Method | Celiac Disease | | | Renal Cell Carcinoma | | |
| --- | --- | --- | --- | --- | --- | --- |
| | Accuracy | F1-score | AUC | Accuracy | F1-score | AUC |
| BYOL | 0.7958 (0.0205) | 0.7750 (0.0286) | 0.9427 (0.0092) | 0.5802 (0.0072) | 0.5334 (0.0151) | 0.8390 (0.0073) |
| BYOL + HistoPerm | **0.8770 (0.0049)** | **0.8773 (0.0062)** | **0.9721 (0.0018)** | **0.6084 (0.0101)** | **0.5604 (0.0055)** | **0.8530 (0.0066)** |
| SimCLR | 0.8507 (0.0031) | 0.8473 (0.0038) | 0.9583 (0.0016) | 0.5920 (0.0069) | 0.5359 (0.0029) | 0.8452 (0.0104) |
| SimCLR + HistoPerm | **0.8855 (0.0057)** | **0.8832 (0.0061)** | **0.9767 (0.0023)** | **0.6033 (0.0121)** | **0.5433 (0.0109)** | **0.8634 (0.0041)** |
| VICReg | 0.7717 (0.0164) | 0.7394 (0.0266) | 0.9218 (0.0073) | 0.5621 (0.0033) | 0.4940 (0.0075) | 0.8074 (0.0068) |
| VICReg + HistoPerm | **0.8501 (0.0092)** | **0.8442 (0.0109)** | **0.9602 (0.0039)** | **0.5890 (0.0005)** | **0.5336 (0.0033)** | **0.8263 (0.0022)** |

**Table 1.** Patch-level linear performance results on the respective test sets. All reported values are the mean with standard deviation in parentheses. The top results for each architecture are presented in **boldface**.

### 4.2 Slide-level Results

In Table 2 we present the effects of HistoPerm on slide-level classification performance. For slide level classification we utilized average-pooling to aggregate the patch-level predictions. This slide-level aggregation approach is straightforward and keeps the evaluation focus on the impact of HistoPerm. Our results showed that incorporating HistoPerm improved performance for all cases of the CD dataset compared to the fully-supervised baseline. On the RCC dataset, the models with HistoPerm showed improved performance for BYOL and VICReg, although all models fell short of the fully-supervised baseline.

|  | Celiac Disease | | | Renal Cell Carcinoma | | |
|---|---|---|---|---|---|---|
| Method | Accuracy | F1-score | AUC | Accuracy | F1-score | AUC |
| Fully-Supervised | 0.9167 (0.0111) | 0.9168 (0.0109) | 0.9856 (0.0005) | 0.7393 (0.0074) | 0.6716 (0.0122) | 0.9608 (0.0071) |
| BYOL | 0.8077 (0.0333) | 0.7918 (0.0477) | 0.9823 (0.0035) | 0.6068 (0.0074) | 0.5274 (0.0190) | 0.9409 (0.0058) |
| BYOL + HistoPerm | **0.9808 (0.0000)** | **0.9804 (0.0000)** | **0.9967 (0.0015)** | **0.6410 (0.0128)** | **0.5661 (0.0111)** | **0.9477 (0.0097)** |
| SimCLR | 0.9423 (0.0192) | 0.9421 (0.0185) | 0.9928 (0.0044) | **0.6410 (0.0256)** | **0.5695 (0.0333)** | 0.9422 (0.0060) |
| SimCLR + HistoPerm | **0.9679 (0.0111)** | **0.9672 (0.0114)** | **0.9961 (0.0028)** | 0.6282 (0.0222) | 0.5331 (0.0319) | **0.9536 (0.0043)** |
| VICReg | 0.7821 (0.0111) | 0.7669 (0.0169) | 0.9823 (0.0030) | 0.5726 (0.0196) | 0.4430 (0.0336) | 0.9196 (0.0094) |
| VICReg + HistoPerm | **0.9423 (0.0192)** | **0.9424 (0.0204)** | **0.9978 (0.0014)** | **0.5897 (0.0000)** | **0.4888 (0.0063)** | **0.9287 (0.0032)** |

**Table 2.** Slide-level linear performance results on the respective test sets. All reported values are the mean of three different runs with standard deviation in parentheses. The top results for each architecture are presented in **boldface**. We provide the supervised results on the top row for comparison.

## 5. DISCUSSION

In this study, we presented HistoPerm, an approach for generating views of histology images to improve representation learning. HistoPerm leverages the weakly-labeled nature of histology images to expand the available pool of views. By expanding the available view pool, we improved the learned representation quality and observed enhanced downstream performance. Our results suggest that HistoPerm is a promising approach for medical image analysis in digital pathology when access to labeled data is limited.

We incorporated HistoPerm into BYOL, SimCLR, and VICReg, and showed improvement in classification performance on two histology datasets. At the patch level, adding HistoPerm to BYOL, SimCLR, and VICReg improved accuracy by 8%, 3%, and 8% on the CD dataset. Similarly, on the RCC dataset, models with HistoPerm outperformed on accuracy by 2%, 1%, and 2% for BYOL, SimCLR, and VICReg, respectively. For CD, we see that models with HistoPerm at the slide-level increase accuracy by 18%, 2%, and 22% for BYOL, SimCLR, and VICReg, respectively. On the RCC data, HistoPerm increases slide-level accuracy by 4% on BYOL and 1% on VICReg, but decreases performance by 2% for SimCLR. Critically, HistoPerm was able to outperform fully-supervised models at the slide-level without patch-level annotations. These findings have important implications for using unlabeled histology images in clinical settings, as image annotation can be a labor-intensive and highly skilled process. Reducing the need for labeled data using HistoPerm, would increase the utility of existing representation learning approaches.

We demonstrated that the addition of HistoPerm can lead to a notable performance improvement on the CD dataset compared to the fully-supervised baseline. However, this trend was not observed on the RCC dataset, where all models performed worse than the fully-supervised baseline. For whole-slide classification, we used an average-pooling approach to aggregate the patch-level predictions. We expect that as we utilize more sophisticated approaches, like multi-head attention or self-attention, our slide-level classification results will outperform the presented results, including the fully-supervised ones.

Of note, the results on the RCC dataset did not show as much improvement as those on the CD dataset. It is possible that this difference is due to the higher morphological complexity and variability of the RCC samples, as indicated by the original study on this dataset.[14] Despite the smaller improvements on the RCC dataset, the use of HistoPerm on both datasets showed clear benefits over standard representation learning approaches. In future work, we plan to investigate the relationship between histological pattern complexity and learned representation quality to enhance the ability of the model to generate more representative features. Furthermore, we intend on expanding our work to explore the biological underpinnings in depth.

While HistoPerm requires less labeled data than fully-supervised approaches, it still requires some labeled data. In future work, we aim to reduce the labeled data requirements further for HistoPerm to enable use in labeled data-constrained settings. We also plan to examine the impact of incorporating unlabeled data from diverse data sources to explore the generalizability of HistoPerm across histology datasets with varied preparation and scanning procedures. In addition, we intend to utilize datasets from multiple disease types and evaluate the effectiveness of the learned histologic representations for transfer learning. This is particularly relevant as data for certain disease types may be scarce, and pretrained representations could provide a solution for building effective image analysis models. Finally, we will explore datasets for tasks like survival prediction in future work.

## 6. CONCLUSION

The presented study showed that the proposed permutation-based view generation method, HistoPerm, offered improved histologic feature representations and resulted in enhanced classification accuracy compared to current representation learning techniques. In some cases, HistoPerm even outperformed the fully-supervised model. This approach allows for the incorporation of unlabeled histology data alongside labeled data for representation learning, resulting in overall higher classification performance. Additionally, the use of HistoPerm may reduce the annotation workload for pathologists, making it a viable option for various digital pathology applications.

**ACKNOWLEDGMENTS**

The authors would like to thank Maxwell Aladago, Wayner Barrios, Yiren Jian, Shuai Jiang, Qingyuan Song, and Yuansheng Xie for their help and suggestions to improve the manuscript. Additionally, the authors would like to thank Naofumi Tomita for his help and suggestions to improve the manuscript and the figures.

**CONFLICT OF INTEREST STATEMENT**

None Declared.

**FUNDING**

This research was supported in part by grants from the US National Library of Medicine (R01LM012837 and R01LM013833) and the US National Cancer Institute (R01CA249758).

# Supplementary Material

## Appendix A. Hyperparameter Search on $\alpha$

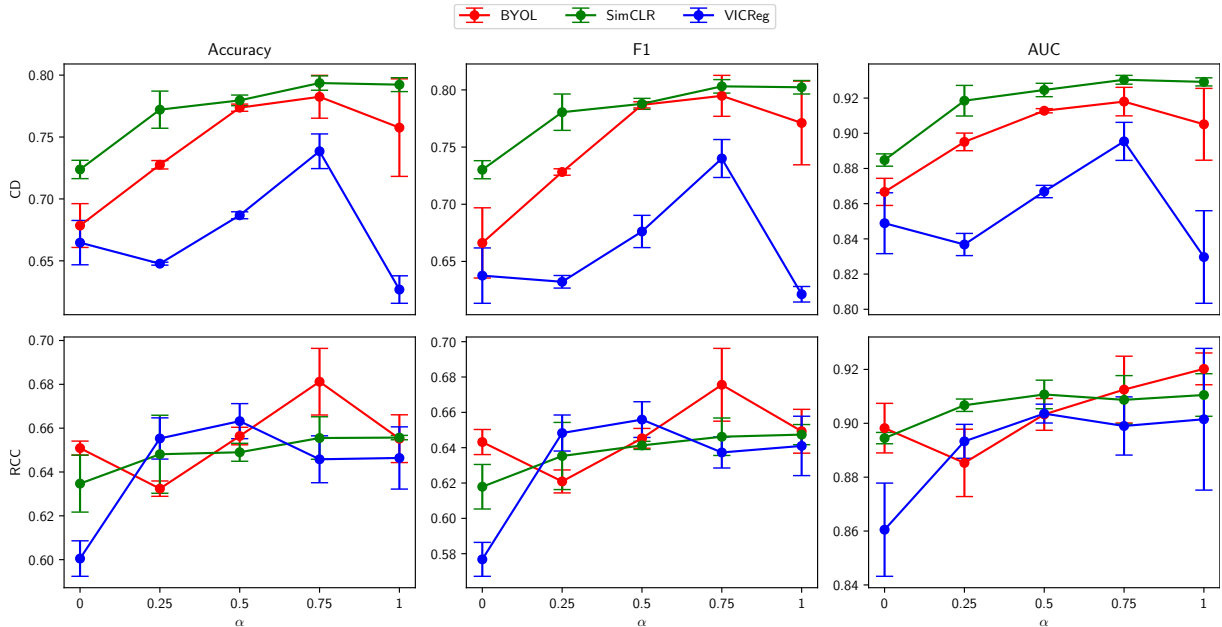

**Supplementary Figure 1.** Patch-level linear evaluation mode results over $\alpha$, the percentage of the mini-batch eligible for permutation. We report all results on the respective development sets.

To determine the optimal value of $\alpha$, we perform a search over $\alpha \in \{0, 0.25, 0.5, 0.75, 1\}$. Of note, $\alpha = 0$ is equivalent to the standard BYOL, SimCLR, and VICReg methods, as all elements of the mini-batch are considered unlabeled. We present our results in Supplementary Figure 1 for the patch-level linear evaluation mode on the respective development sets. From our experiments over $\alpha$, we conclude that selecting $\alpha = 0.75$ is a good choice for both datasets, as in all cases, it outperforms the standard methods in terms of accuracy. We find that almost any value of $\alpha$ results in improved performance over the respective baseline methods in terms of accuracy, indicating that most permutation choices should provide a measurable improvement. Lastly, even suboptimal values of $\alpha$ produce promising results, so expensive tuning of this hyperparameter may not be necessary in most cases.

# Appendix B. Data Transformation Hyperparameter Search

We performed a hyperparameter search on the applied image augmentations for use in the pretraining phase. The tested image augmentations are as follows:

- **Cropping:** Randomly sample a crop from the image between 8% and 100% of the original image size (224×224 pixels for our datasets). Additionally, randomly sample an aspect ratio between 3/4 and 4/3. Resize images to 224×224 pixels using bilinear interpolation.
- **Flipping:** Flip the images over the horizontal and vertical axes. Flip over both axes, as histology images are rotation invariant.
- **Color Jittering:** Randomly change the brightness, contrast, hue, and saturation of the image according to a value uniformly selected from a range.
- **Grayscale:** Convert the image to grayscale. Given intensities $(r, g, b)$ for a pixel in the image, convert it to grayscale according to the formula: $0.299r + 0.587g + 0.114b$.
- **Gaussian Blurring:** Use a 23×23-pixel Gaussian kernel to apply blurring with the standard deviation sampled from the range [0.1, 2.0].
- **Solarization:** Invert all pixels in the image above a threshold.

We use the Kornia computer vision deep learning library for all augmentations[1]. In Supplementary Table 1, we provide the hyperparameters for the tested data transformations.

| Hyperparameter | $\mathcal{T}_1$ | $\mathcal{T}_2$ |
|---|---|---|
| Cropping probability | 1.0 | 1.0 |
| Flipping probability | 0.5 | 0.5 |
| Color jittering | | |
|     probability | 0.8 | 0.8 |
|     brightness factor | 0.4 | 0.4 |
|     contrast factor | 0.4 | 0.4 |
|     hue factor | 0.1 | 0.1 |
|     saturation factor | 0.2 | 0.2 |
| Grayscale probability | 0.2 | 0.2 |
| Gaussian blurring probability | 1.0 | 0.1 |
| Solarization | | |
|     probability | 0.0 | 0.2 |
|     threshold | 128/255 | 128/255 |

**Supplementary Table 1.** Hyperparameter settings for tested data transformations.

## Appendix B.1 Data Transformation Set

We detail the tested data transformation sets as follows:

- **Base:** Apply all transformations as enumerated in Supplementary Table 1.

- **Remove Grayscale:** Apply all transformations as enumerated in Supplementary Table 1, apart from random conversion to grayscale.

- **Remove Color:** Apply all transformations as enumerated in Supplementary Table 1, apart from random conversion to grayscale, color jittering, and solarization.

- **Crop + Blur + Flipping:** Apply cropping, Gaussian blurring, and flipping as enumerated in Supplementary Table 1.

- **Crop + Flipping:** Apply cropping and flipping as enumerated in Supplementary Table 1.

- **Blur + Flipping:** Apply Gaussian blurring and flipping as enumerated in Supplementary Table 1.

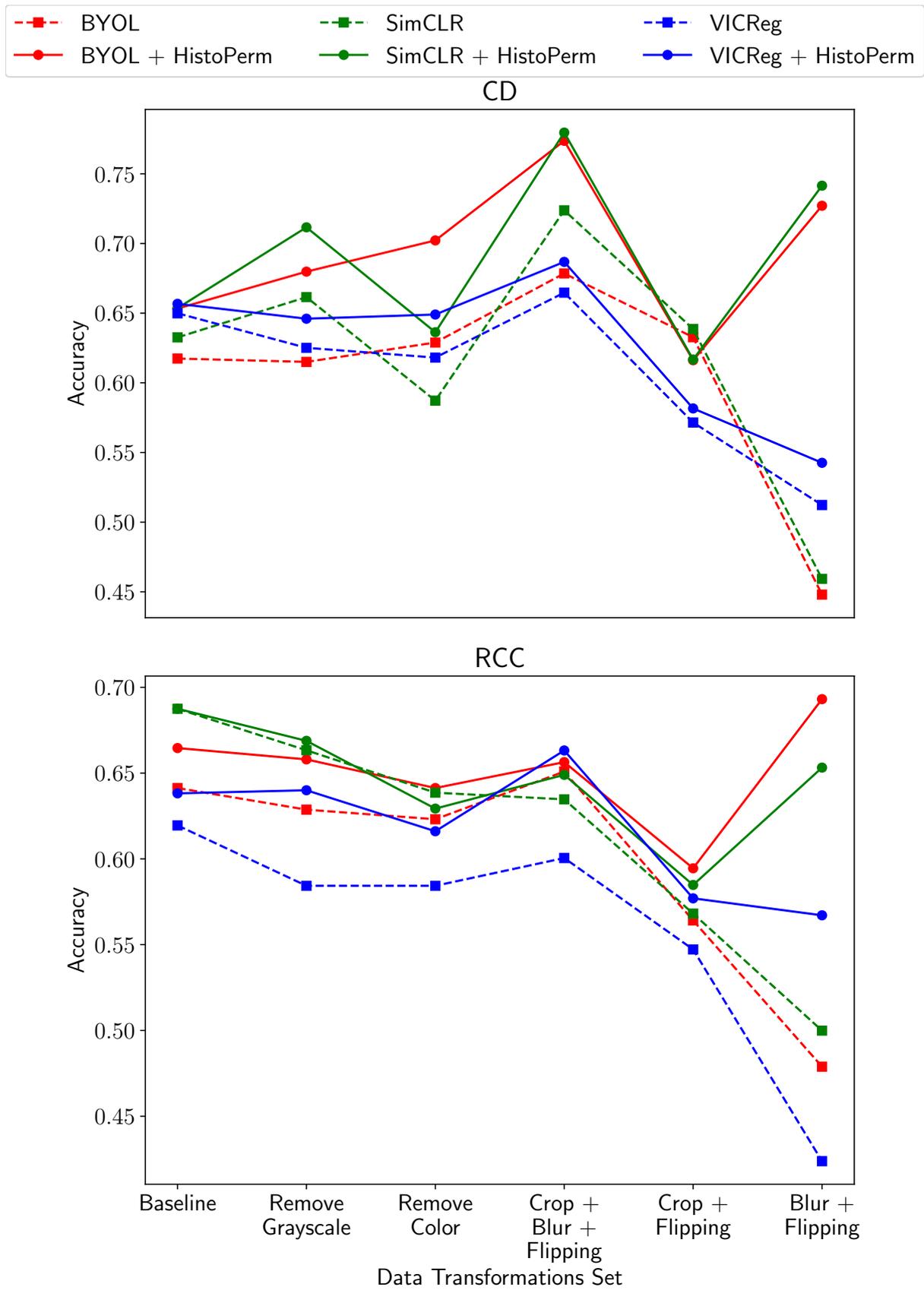

**Supplementary Figure 2.** Linear performance results on the CD and RCC development sets for the tested data transformations.

In Supplementary Figure 2, we present the results of our hyperparameter search on the data transformations. Overall, we find that adding HistoPerm provides a measurable classification accuracy improvement in all cases apart from "Crop + Flipping" on the CD dataset. We select "Crop + Blur + Flipping" as the best performing set of data transformations, because it performs well both with and without HistoPerm and should provide the most equitable base for demonstrating the benefits of our approach. Using only "Blur + Flipping" does result in overall higher classification accuracy with HistoPerm, except the results without HistoPerm are significantly lower and are not representative of a baseline model.

**Appendix C. Architecture Details**

In a joint embedding architecture, there are two or three main components depending on the method being used. BYOL consists of an encoder, projector, and predictor, SimCLR uses an encoder and projector, and VICReg uses an encoder and expander. We discuss the structure of each component below.

1. **Encoder:** ResNet18[2] feature extractor (i.e., the output of the final average pooling layer), $f$, producing representation $y$ such that $y = f(x) \in \mathbb{R}^{\mathcal{D}_f}$. We use $\mathcal{D}_f = 512$ in all experiments.

2. **Projector:** Multilayer perceptron, $g$, mapping $y$ to $z = g(y) \in \mathbb{R}^{\mathcal{D}_g}$. We instantiate $g$ as a two-layer multilayer perceptron with a single hidden layer of size 4096 and output size $\mathcal{D}_g = 256$.

3. **Expander:** Multilayer perceptron, $h$, mapping $y$ to $z = h(y) \in \mathbb{R}^{\mathcal{D}_h}$. We instantiate $h$ as a two-layer multilayer perceptron with a single hidden layer of size 2048 and output size $\mathcal{D}_h = 2048$.

4. **Predictor:** Multilayer perceptron, $q$, mapping $z$ to $p = q(z) \in \mathbb{R}^{\mathcal{D}_q}$. We instantiate $q$ as a two-layer multilayer perceptron with a single hidden layer of size 4096 and output size $\mathcal{D}_q = 256$.

Now, we cover how these components are used in the standard BYOL, SimCLR, and VICReg methods as well as when HistoPerm is added.

**Appendix C.1 BYOL**

The BYOL architecture is split across two components called online and target branches, parameterized by $\theta$ and $\xi$, respectively. The online branch is composed of three stages: encoder $f_\theta$, projector $g_\theta$, and predictor $q_\theta$. Likewise, the target branch has two stages: encoder $f_\xi$ and projector $g_\xi$. Encoders $f_\theta$ and $f_\xi$ map input views to a representation space, which are then fed to respective projectors $g_\theta$ and $g_\xi$. Note that the predictor $q_\theta$ is only used in the online network, as prior works have shown that this architectural asymmetry is necessary to avoid collapsing to the trivial solution.[3] At the end of training, we only keep the online encoder $f_\theta$ and use the pretrained weights as initialization for the fully-supervised downstream tasks.

Learning progresses by computing the mean squared error between online and target branches. For both unlabeled and labeled views, we compute the loss as follows:

$$\mathcal{L}_{BYOL} = \underbrace{\|q_\theta(z_{u,1}) - \text{sg}(z_{u,2})\|}_{Loss_u} + \underbrace{\|q_\theta(z_{l,1}) - \text{sg}(\widetilde{z_{l,2}})\|}_{Loss_l}$$

where $\text{sg}(\cdot)$ is the stop-gradient operation, so the target branch weights are not updated through optimization. As shown in the SimSiam paper[3], the stop-gradient is necessary to avoid collapse. We symmetrize the loss by passing $v_{u,2}$, $\widetilde{v_{l,2}}$ and $v_{u,1}$, $v_{l,1}$ to the online and target branches, respectively. In the code implementation for the loss function, we combine and process both mini-batches simultaneously to avoid inadvertently leaking any information about the source dataset through batch normalization. Note that in the case where $v_{l,1} = \widetilde{v_{l,2}} = \emptyset$, (i.e., $\mathcal{P}_l = \mathcal{P}_u = \emptyset$), our method reduces to the default BYOL formulation. Then, we perform the weight updates as follows:

$$\theta \leftarrow \text{optimizer}(\theta, \nabla_\theta \mathcal{L}, \eta)$$

$$\xi \leftarrow \tau\xi + (1-\tau)\xi$$

where $\tau \in [0, 1]$ is a momentum hyperparameter and $\eta$ is the learning rate for gradient descent.

**Appendix C.2 SimCLR**

The SimCLR architecture consists of an encoder $f_\phi$ and projector $g_\phi$ parameterized by weights $\phi$. The encoder $f_\phi$ maps pairs of input views to representations that are then fed to the projector $g_\phi$. Finally, $L_2$-normalization is applied to the outputs of the projector. At the end of training, we only keep the encoder $f_\phi$ and use the pretrained weights as initialization for the fully-supervised downstream tasks.

Learning progresses using the NT-Xent loss as defined in the original SimCLR formulation.[4] For both unlabeled and labeled views, we compute the loss as follows:

$$\mathcal{L}_{SimCLR} = \frac{1}{|\mathcal{X}_u| + |\mathcal{X}_l|} \left[ \underbrace{\sum_{i=1}^{|\mathcal{X}_u|} -\log \frac{\exp(z_i \cdot z_i'/\tau)}{\sum_{k=1}^{|\mathcal{X}_u|+|\mathcal{X}_l|} \exp(z_i \cdot z_k/\tau)}}_{Loss_u} \right.$$

$$\left. + \underbrace{\sum_{i=|\mathcal{X}_u|+1}^{|\mathcal{X}_u|+|\mathcal{X}_l|} -\log \frac{\exp(z_i \cdot z_i'/\tau)}{\sum_{k=1}^{|\mathcal{X}_u|+|\mathcal{X}_l|} \exp(z_i \cdot z_k/\tau)}}_{Loss_l} \right]$$

where $\tau$ is the temperature hyperparameter. Note that in the case where $\widetilde{v_{l,1}} = v_{l,2} = \emptyset$ (i.e., $\mathcal{P}_\ell = \mathcal{X}_\ell = \emptyset$), our method reduces to the default SimCLR formulation.

Then, we perform the weight updates as follows:

$$\phi \leftarrow \text{optimizer}(\phi, \nabla_\phi \mathcal{L}, \eta)$$

where $\eta$ is the learning rate for gradient descent.

## Appendix C.3 VICReg

The VICReg architecture consists of an encoder $f_\psi$ and expander $h_\psi$ parametrized by weights $\psi$. The encoder $f_\psi$ maps pairs of input views to representations that are then fed to the expander $h_\psi$. At the end of training, we only keep the encoder $f_\psi$ and use the pretrained weights as initialization for the fully-supervised downstream tasks.

Learning progresses using variance, invariance, and covariance loss terms.[5] The variance term is as follows:

$$v(Z) = \frac{1}{d} \sum_{j=1}^{d} \max\left(0, \gamma - S(z^j, \epsilon)\right)$$

where

$$S(x, \epsilon) = \sqrt{\text{Var}(x) + \epsilon}$$

Now, the invariance term is:

$$s = \underbrace{\frac{1}{|\mathcal{X}_u|} \sum_{i=1}^{|\mathcal{X}_u|} \|z_i - z_2'\|_2^2}_{s_u} + \underbrace{\frac{1}{|\mathcal{X}_l|} \sum_{i=|\mathcal{X}_u|+1}^{|\mathcal{X}_u|+|\mathcal{X}_l|} \|z_i - z_i'\|_2^2}_{s_l}$$

Lastly, the covariance term is:

$$C(Z) = \frac{1}{|\mathcal{X}_u| + |\mathcal{X}_l| - 1} \left( \underbrace{\sum_{i=1}^{|\mathcal{X}_u|} (z_i - \bar{z})(z_i - \bar{z})^T}_{C_u} + \underbrace{\sum_{i=|\mathcal{X}_u|+1}^{|\mathcal{X}_u|+|\mathcal{X}_l|} (z_i - \bar{z})(z_i - \bar{z})^T}_{C_l} \right)$$

where

$$\bar{z} = \frac{1}{|\mathcal{X}_u| + |\mathcal{X}_l|} \left( \underbrace{\sum_{i=1}^{|\mathcal{X}_u|} z_i}_{\bar{z}_u} + \underbrace{\sum_{i=|\mathcal{X}_u|+1}^{|\mathcal{X}_u|+|\mathcal{X}_l|} z_i}_{\bar{z}_l} \right)$$

Given all these loss terms, the overall loss function is now:

$$\mathcal{L}_{VICReg} = \lambda s(Z, Z') + \mu[v(Z) + v(Z')] + \nu[c(Z) + c(Z')]$$

where $\lambda$, $\mu$, and $\nu$ are hyperparameters to weight the effect of each loss term. Note that in the case where $\widetilde{v_{l,1}} = v_{l,2} = \emptyset$ (i.e., $\mathcal{P}_\ell = \mathcal{X}_\ell = \emptyset$), our method reduces to the default VICReg formulation.

## Appendix D. Implementation Details

In this section, we provide all needed implementation details and hyperparameters for reproducing our model.

### Appendix D.1. Pre-training

- **Cropping:**
    - Size: 224×224 pixels
    - Scale range: (0.08, 1.0)
    - Ratio range: (3/4, 4/3)
- **Encoder:** ResNet18[2]
- **Epochs:** 50
- $\eta$: $10^{-3}$
- **Gaussian blurring:**
    - Kernel size: 23×23 pixels
    - Standard deviation range: [0.1, 2.0]
- **Learning rate:** 0.45
- **Learning rate scheduler:** Cosine decay[6]
- **Learning rate warm-up epochs:** 5

- **Mini-batch size:** 256
- **Momentum:** 0.9
- **Optimizer:** LARS[7]
- **View 1 augmentations:**
  - Cropping probability: 1.0
  - Flipping probability: 0.5
  - Gaussian blurring probability: 1.0
- **View 2 augmentations:**
  - Cropping probability: 1.0
  - Flipping probability: 0.5
  - Gaussian blurring probability: 0.1
- **Weight decay:** $10^{-6}$
- **BYOL-specific:**
  - Projectors $g_\theta$ and $g_\xi$:
    - Number of layers: 2
    - Hidden dimension: 4096
    - Output dimension: 256
  - Predictor $q_\theta$:
    - Number of layers: 2
    - Hidden dimension: 4096
    - Output dimension: 256
  - $\tau$: 0.97
- **SimCLR-specific:**
  - Projector $g_\phi$:
    - Number of layers: 2

- Hidden dimension: 4096
- Output dimension: 256
    - $\tau$: 1.0
- **VICReg-specific:**
    - Expander $h_\psi$:
        - Number of layers: 2
        - Hidden dimension: 2048
        - Output dimension: 2048
    - $\mu$: 25
    - $\lambda$: 25
    - $\nu$: 1
    - $\epsilon$: $10^{-4}$

## Appendix D.2. Linear

- **Cropping:**
    - Size: 224×224 pixels
    - Scale range: (0.08, 1.0)
    - Ratio range: (3/4, 4/3)
- **Data augmentations:**
    - Cropping probability: 1.0
    - Flipping probability: 0.5
- **Epochs:** 80
- **Learning rate:** 0.2
- **Learning rate scheduler:** Cosine decay[6]
- **Learning rate warm-up epochs:** 5
- **Mini-batch size:** 256

- **Optimizer:** SGD with Nesterov mometum[8]

## Appendix D.3. Fully-Supervised

- **Color jittering:**
    - Brightness factor: 0.5
    - Contrast factor: 0.5
    - Hue factor: 0.2
    - Saturation factor: 0.5
- **Data augmentations:**
    - Color jittering probability: 1.0
    - Flipping probability: 1.0
    - Rotation probability: 1.0
- **Epochs:** 40
- **Learning rate:** $10^{-4}$
- **Learning rate scheduler:** Decay by a factor of 0.85 each epoch
- **Mini-batch size:** 16
- **Optimizer:** Adam[9]
- **Model:** ResNet18[2]
- **Weight decay:** $10^{-4}$

# Appendix E. Dataset Statistics

## Appendix E.1. Celiac Disease Dataset

In Supplementary Tables 2 and 3 we present the dataset distribution for the CD dataset at the patch- and slide-level, respectively.

| Class | Training | Development | Testing |
|---|---|---|---|
| Normal | 81,428 | 4,535 | 5,966 |
| Nonspecific duodenitis | 80,090 | 9,049 | 10,010 |
| Celiac sprue | 81,560 | 9,593 | 10,038 |
| Total | 243,078 | 23,177 | 25,683 |

**Supplementary Table 2.** The Celiac disease patch-level dataset splits for the Normal, Nonspecific duodenitis, and Celiac sprue classes. The assigned label per patch is the corresponding slide-level label.

| Class | Training | Development | Testing |
|---|---|---|---|
| Normal | 75 | 19 | 16 |
| Nonspecific duodenitis | 83 | 16 | 18 |
| Celiac sprue | 93 | 19 | 18 |
| Total | 251 | 54 | 52 |

**Supplementary Table 3.** The Celiac disease slide-level dataset splits for the Normal, Nonspecific duodenitis, and Celiac sprue classes.

## Appendix E.2. Renal Cell Carcinoma Dataset

In Supplementary Tables 4 and 5 we present the dataset distribution for the RCC dataset at the patch- and slide-level, respectively.

| Class | Training | Development | Testing |
|---|---|---|---|
| Benign | 80,165 | 1,963 | 4,372 |
| Oncocytoma | 80,903 | 1,457 | 5,470 |
| Chromophobe | 80,660 | 3,279 | 10,387 |
| Clear cell | 78,355 | 3,059 | 12,446 |
| Papillary | 83,867 | 2,259 | 9,987 |
| Total | 403,950 | 12,017 | 42,662 |

**Supplementary Table 4.** The Renal Cell Carcinoma patch-level dataset splits for the Benign, Oncocytoma, Chromophobe, Clear cell, and Papillary classes. The assigned label per patch is the corresponding slide-level label.

| Class | Training | Development | Testing |
|---|---|---|---|
| Benign | 14 | 5 | 10 |
| Oncocytoma | 14 | 3 | 10 |
| Chromophobe | 15 | 5 | 18 |
| Clear cell | 285 | 5 | 20 |
| Papillary | 55 | 5 | 20 |
| Total | 383 | 23 | 78 |

**Supplementary Table 5.** The Renal Cell Carcinoma slide-level dataset splits for the Benign, Oncocytoma, Chromophobe, Clear cell, and Papillary classes.